\documentclass[a4paper, 10 pt, conference]{ieeeconf}  

\IEEEoverridecommandlockouts                              

\usepackage{amsmath} 
\usepackage{amssymb}  
\usepackage{lipsum}
\usepackage{color}
\usepackage[normalem]{ulem}

\usepackage{graphicx}
\usepackage{bm}
\usepackage{microtype}
\usepackage{threeparttable}
\usepackage{booktabs}
\usepackage{multirow}

\usepackage[none]{hyphenat}
\title{\LARGE \bf
Reference-Augmented Learning for Precise Tracking Policy of Tendon-Driven Continuum Robots
}

\author{Ziqing Zou, Ke Qiu, Haojian Lu, Rong Xiong and Yue Wang$^*$
\thanks{$^{*}$Corresponding author (e-mail: ywang24@zju.edu.cn).}%
\thanks{
All the authors are with Department of Control Science and Engineering, Zhejiang University, Hangzhou, China.}%
}

\begin{document}
\maketitle
\thispagestyle{empty}
\pagestyle{empty}

\begin{abstract}
Tendon-Driven Continuum Robots (TDCRs) pose significant control challenges due to their highly nonlinear, path-dependent dynamics and non-Markovian characteristics. Traditional Jacobian-based controllers often struggle with hysteresis-induced oscillations, while conventional learning-based approaches suffer from poor generalization to out-of-distribution trajectories. This paper proposes a reference-augmented offline learning framework for precise 6-DOF tracking control of TDCRs. By leveraging a differentiable RNN-based dynamics surrogate as a gradient bridge, we optimize a control policy through an augmented reference distribution. This multi-scale augmentation scheme incorporates stochastic bias, harmonic perturbations, and random walks, forcing the policy to internalize diverse tracking error recovery mechanisms without additional hardware interaction. Experimental results on a three-section TDCR platform demonstrate that the proposed policy achieves a 50.9\% reduction in average position error compared to non-augmented baselines and significantly outperforms Jacobian-based methods in both precision and stability across various speeds.
For implementation details and source code, please refer to https://github.com/ZiqingZou/ContinuumControl.
\end{abstract}

\section{Introduction}

Tendon-Driven Continuum Robots (TDCRs) have attracted significant attention in minimally invasive surgery~\cite{burgner2015survey}, in-situ aeroengine inspection~\cite{qi2026inspection}, and dexterous manipulation~\cite{Camarillo2008manipulators} due to their inherent compliance and high maneuverability in constrained environments.
Unlike rigid-link robots, TDCRs rely on the coordinated contraction of internal tendons to deform a flexible backbone. However, achieving precise and agile motion control remains a formidable challenge. The complex interplay between the viscoelastic deformation of the backbone and the nonlinear friction along the tendon routing paths introduces significant hysteresis and dead-zone effects~\cite{shen2026friction}. These phenomena render the system dynamics inherently non-Markovian, where the current robot state is deeply coupled with its actuation history, leading to path-dependent configuration deviations and significant response latencies.

Traditional control strategies for TDCRs predominantly rely on Jacobian-based kinematic schemes or simplified static models~\cite{qiu2025actuator, Fang2019gaussian}. While these methods are computationally efficient, they often treat the robot as a quasi-static geometric entity, failing to account for high-order dynamic characteristics~\cite{kazemipour2022adaptive}. Consequently, under high-speed tracking or complex trajectory commands, Jacobian-based controllers are prone to self-excited oscillations and persistent tracking offsets. On the other hand, learning-based approaches, such as supervised learning or imitation learning, have shown promise in capturing nonlinearities~\cite{hendrik2024rnn, thuruthel2017learning, tim2026pinn}. However, these methods typically suffer from poor generalization. Policies trained on smooth, pre-defined demonstration trajectories often fail when encountering out-of-distribution (OOD) references or sharp command transitions, limiting their reliability in dynamic real-world tasks.

Reinforcement Learning (RL) can alleviate generalization issues to a certain extent~\cite{thuruthel2019model, satheeshbabu2019open, Centurelli2022close}. Through well-designed reference trajectories and exhaustive exploration of the state space, RL agents can mitigate OOD sensitivity and asymptotically converge to an optimal tracking policy~\cite{jemin2019hutter}. However, its primary drawback lies in low sample efficiency, necessitating extensive interactions with the environment and posing inherent safety risks during the data collection phase. Although Goal-Conditioned Reinforcement Learning (GCRL) improves data utilization by leveraging negative samples~\cite{wang2025thousand}, it still fails to fundamentally resolve the prohibitive interaction demands characteristic of online RL.

To address these limitations, this paper proposes a reference-augmented offline learning framework for the precise and robust tracking control of TDCRs. The core intuition is that a robust policy must internalize the system's long-term temporal dependencies and diverse distributional shifts during the training phase. Leveraging a differentiable RNN-based dynamics model as a gradient bridge~\cite{zou2025excavator}, we optimize the control policy by backpropagating tracking errors through an augmented reference distribution. Specifically, instead of following fixed trajectories, we dynamically generate randomized reference signals comprising sinusoidal oscillations and step functions, centered around the robot’s subsequent states. The sinusoidal components force the policy to compensate for phase lags and hysteresis loops, while the step components sharpen the model's ability to handle sudden setpoint changes without inducing oscillations. This approach allows the policy to explore a much denser manifold of the state-action space without requiring additional physical hardware interaction.

The primary contributions of this work are summarized as follows:
\begin{itemize}
    \item \textbf{Reference-Augmented Learning Framework}: We introduce a novel policy training pipeline that facilitates the learning of the precise tracking policy through reference trajectory randomization.
    \item \textbf{Differentiable Policy Optimization}: We utilize a high-fidelity RNN-based dynamics surrogate to enable end-to-end gradient flow, allowing the policy to implicitly internalize complex compensation laws for friction and compliance, thereby enhancing tracking performance.
    \item \textbf{Experimental Validation}: We demonstrate on a physical three-section TDCR platform that the proposed policy significantly outperforms conventional Jacobian-based methods.
\end{itemize}

\begin{figure*}[t]
    \vspace{5pt}
    \centering
    \includegraphics[width=6.8in]{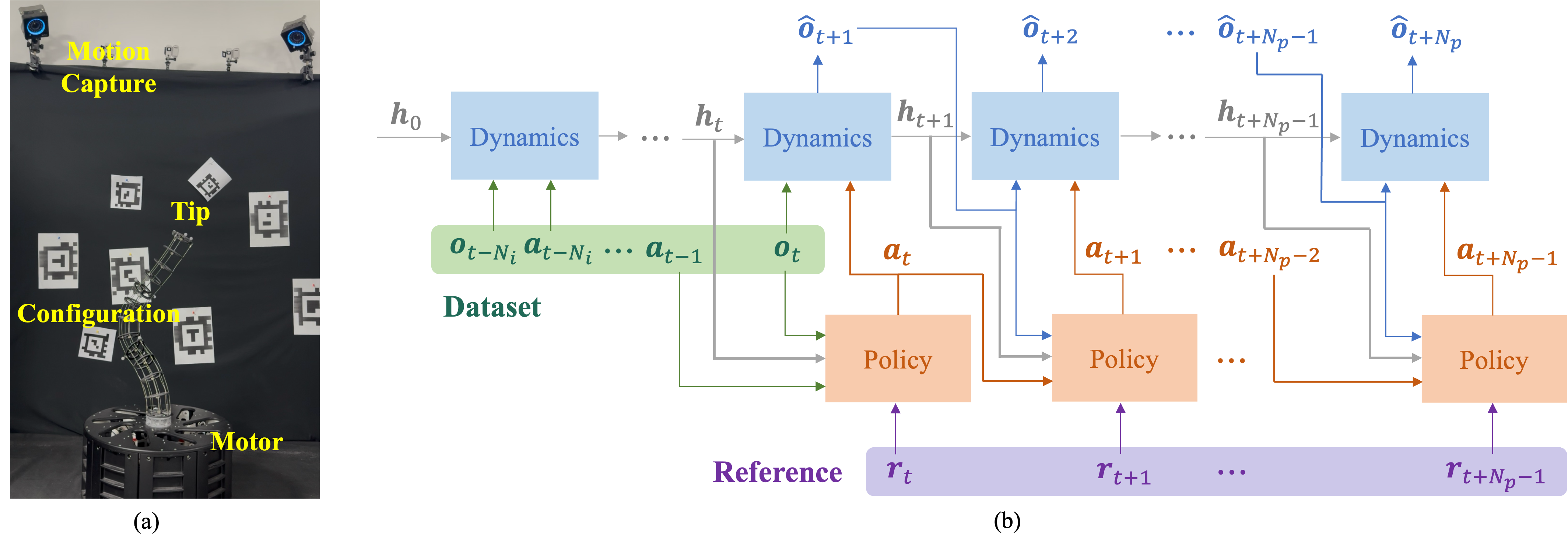}
    \vspace{-12pt}
    \caption{(a) The three-section TDCR experimental platform. The robot is actuated by nine motors that independently control tendon displacements to modulate its configuration and tip pose. A motion capture system is mounted above the workspace to provide high-precision ground-truth measurements of the robot's tip position and orientation. (b) Training pipeline of the neural tracking policy. Historical trajectories are sampled from the dataset (green) to initialize the hidden state via the dynamics model (gray). Conditioned on the look-ahead reference (purple), the policy network generates control actions (orange), which are then fed into the dynamics surrogate to predict future observations (blue) and update the hidden state (gray) in an auto-regressive manner.}
    \label{fig:total}
    \vspace{-5pt}
\end{figure*}

\section{Model Settings}

\subsection{Dynamics Surrogate}
The precision control of Tendon-Driven Continuum Robots (TDCRs) is fundamentally constrained by their complex structural dynamics. Conventional modeling approaches typically rely on constant curvature (PCC) kinematics~\cite{webster2010design} or static mechanics, which often overlook the inherent time-varying and history-dependent behaviors of the system. In practice, the mapping from motor actuation to tip pose is perturbed by significant transmission nonlinearities, including viscoelastic deformation of the backbone and frictional hysteresis along the tendon routing paths.

To address these challenges, we represent the TDCR as a high-order dynamic system. Unlike memory-less geometric models, our approach employs a high-fidelity RNN-based dynamics surrogate to capture the temporal dependencies. The core advantage of this model lies in its recurrent hidden state $\bm{h}_t$, which serves as a latent encoding of the robot's physical context. By formulating the dynamics in a residual manner, the model predicts the observation evolution as:
\begin{equation}
    \label{eq:dynamics}
    \left\{
    \begin{aligned}
        \bm{h}_{t+1} &= \bm{f}_{\theta}(\bm{h}_t, \bm{o}_t, \bm{a}_t), \\
        \hat{\bm{o}}_{t+1} &= \bm{o}_t + \bm{g}_{\psi}(\bm{h}_{t+1}),
    \end{aligned}
    \right.
\end{equation}
where 
\begin{equation}
    \bm{o}_t = (\bm{l}_t, \bm{v}_t, \bm{T}_t)
\end{equation}
denotes the multi-modal observation vector, comprising the motor rotation lengths $\bm{l}_t$, motor velocities $\bm{v}_t$, and the tip pose. We parameterize the tip pose 
\begin{equation}
    \bm{T} = \left( \bm{p}_t,  \bm{\phi}_t\right) \in \mathbb{R}^6
\end{equation}
with $\bm{p}_t \in \mathbb{R}^3$ and $\bm{\phi}_t \in \mathbb{R}^3$ representing Cartesian position and orientation, respectively.
This differentiable framework allows the dynamics surrogate to serve as a high-fidelity simulator that propagates gradients back to the controller, enabling end-to-end policy optimization.

The dynamics surrogate is trained offline using a large-scale dataset of collected across multiple operational sessions to capture long-term system drift. To further enhance the model's robustness and generalization, stochastic perturbations are injected into the motor commands during the data collection process.

\subsection{Tracking Policy}

The goal of trajectory tracking is to find a control law that minimizes the deviation between the robot's tip pose and a time-varying reference $\bm{T}^*_t \in \mathbb{R}^6$ over a finite horizon. Traditional Jacobian-based control laws~\cite{qiu2025actuator}, typically formulated as:
\begin{equation}
    \bm{a}_t = \mathbf{J}^{\dagger}\left(\bm{T}_t\right) \left( \bm{\dot{T}}^*_t + \mathbf{K}(\bm{T}^*_t - \bm{T}_t )\right), 
\end{equation}
where $\mathbf{J}^{\dagger}$ denotes the pseudo-inverse of the Jacobian matrix computed by weighted damped least-squares method~\cite{4075580}.
Such schemes fundamentally assume a quasi-static relationship between control input and tip pose output, which often treat the Jacobian as a state-dependent but path-invariant operator. Such first-order Markovian assumptions often fail to account for the high-order dynamics inherent in TDCRs, where the system output is deeply coupled with its actuation history.

To overcome these limitations, we leverage the latent hidden states $\bm{h}_t$ of the dynamics model to encapsulate the robot’s temporal context.
Specifically, we formulate the tracking task as a policy optimization problem within a differentiable pipeline. We define a neural control policy $\bm{\pi}_\mu$ that generates recursive action updates $\Delta \bm{a}_t$ by conditioning on the observation $\bm{o}_t$, the latent state $\bm{h}_t$, the previous action $\bm{a}_{t-1}$, and the look-ahead reference $\bm{r}_t$:
\begin{equation}
    \Delta \bm{a}_t = \bm{\pi}_\mu \left(\bm{o}_t, \bm{h}_t, \bm{a}_{t-1}, \bm{r}_t \right),
\end{equation}
where 
\begin{equation}
    \bm{r}_t = \left(\bm{T}^*_{t+1}, \bm{T}^*_{t+2}, \dots, \bm{T}^*_{t+N_r}\right)
\end{equation}
represents the concatenated reference points for $N_r$ future steps, providing the policy with anticipatory capabilities to compensate for system latency.

The objective is to minimize a multi-objective cost function $\mathcal{J}(\mu)$ that balances tracking accuracy and control smoothness:
\begin{equation}
\label{eq:optimization}
    \min_{\mu} \mathcal{J}(\mu) = \mathbb{E}_{\tau \sim (\bm{\pi}_\mu, \bm{f}_\theta, \bm{g}_\psi)} \bigg[ \sum_{i=1}^{N_p} \lambda^{i-1} l_{t+i} \bigg],
\end{equation} 
where 
\begin{equation}
    \tau = \{ (\hat{\bm{o}}_{t+i}, \bm{a}_{t+i}, \bm{h}_{t+i}) \}_{i=1}^{N_p}    
\end{equation}
denotes the trajectory rollout generated by the recursive interaction between the policy $\bm{\pi}_\mu$ and the dynamics surrogate $(\bm{f}_\theta, \bm{g}_\psi)$ over an optimization horizon $N_p$, $\lambda \in \left(0, 1\right]$ is a discount factor.
Here, the step-wise cost is defined as 
\begin{equation}
    \begin{split}
        l_{t+i} &= l \left( \hat{\bm{o}}_{t+i}, \bm{a}_{t+i-1}, \bm{r}_{t+i} \right) \\
        &= \|\hat{\bm{T}}_{t+i} - \bm{T}^*_{t+i} \|^2 + \alpha \|\Delta \bm{a}_{t+i-1}\|^2,
    \end{split}
\end{equation}
where $\alpha$ is a weighting coefficient that penalizes aggressive control effort. 

As illustrated in Fig.~\ref{fig:total}(b), the training process follows a recursive rollout scheme. Each optimization iteration begins by sampling a historical trajectory segment from the dataset $\mathcal{D}$. To ensure that the latent state $\bm{h}_t$ accurately encapsulates the robot’s physical context, we perform a warm-up phase where $N_i$ historical observation-action pairs are recursively processed through the dynamics transition function $\bm{f}_\theta$.
Once initialized, the policy enters an auto-regressive prediction phase for the optimization horizon $N_p$, where the total gradient is computed via Backpropagation Through Time (BPTT)~\cite{zhang2025flight}:

\vspace{-8pt}
\begin{equation}
\label{eq:gradient_flow}
\small
    \begin{split}
        \nabla_\mu \mathcal{J} = & \frac{1}{N_p} \sum_{i=1}^{N_p} \lambda^{i-1} \bigg(
        \frac{\partial l_{t+i}}{\partial \bm{a}_{t+i-1}} \frac{\partial \bm{a}_{t+i-1}}{\partial \mu} \\
        &+ \sum_{j=1}^{i} \frac{\partial l_{t+i}}{\partial \hat{\bm{o}}_{t+i}} \prod_{k=j+1}^{i} \left( \frac{\partial{\hat{\bm{o}}_{t+k}}}{\partial{\hat{\bm{o}}_{t+k-1}}} \right)\frac{\partial \hat{\bm{o}}_{t+j}}{\partial \bm{a}_{t+j-1}} \frac{\partial \bm{a}_{t+j-1}}{\partial \mu} \bigg).
    \end{split}
\end{equation}

Unlike standard reinforcement learning, which relies on high-variance gradient estimation, our framework enables the direct computation of analytical gradients by unrolling this differentiable chain. The total gradient $\nabla_\mu \mathcal{J}$ accounts for both the instantaneous control penalty and the long-term impact of actions on future robot states. This is achieved by backpropagating the tracking errors through the unrolled temporal sequence of the dynamics surrogate, effectively using the model's Jacobians $\frac{\partial{\hat{\bm{o}}_{t+k}}}{\partial{\hat{\bm{o}}_{t+k-1}}}$ as a gradient bridge to refine the control law.

\begin{table*}[t]
\centering
\caption{Tracking Precision Comparison (Position Error in mm, Orientation Error in \textdegree)}
\label{tab:precision_comparison}
\renewcommand{\arraystretch}{1.2}
\begin{tabular}{lcccccccc}
\toprule
\multirow{2}{*}{Method} & \multicolumn{2}{c}{Speed 1.0$\times$} & \multicolumn{2}{c}{Speed 1.7$\times$} & \multicolumn{2}{c}{Speed 2.5$\times$} & \multicolumn{2}{c}{Step Average} \\ \cmidrule(lr){2-3} \cmidrule(lr){4-5} \cmidrule(lr){6-7} \cmidrule(lr){8-9}
& Pos. (mm) & Ori. (\textdegree) & Pos. (mm) & Ori. (\textdegree) & Pos. (mm) & Ori. (\textdegree) & Pos. (mm) & Ori. (\textdegree) \\ \midrule
\textit{Ours (w/ Augmentation)} & \textbf{12.49} & \textbf{4.9} & \textbf{14.04} & \textbf{6.4} & \textbf{18.96} & \textbf{6.9} & \textbf{14.25} & \textbf{5.8} \\
\textit{Dataset Ref (w/o Aug.)} & 25.32 & 7.2 & 30.39 & 8.2 & 36.10 & 10.0 & 29.00 & 8.1 \\
\textit{Jacobian-based} & 24.46 & 23.1 & 32.97 & 21.9 & 43.61 & 20.9 & 30.84 & 22.3 \\ \midrule
\textit{Ours (Pos-only)} & 9.79 & - & 13.41 & - & 17.27 & - & 12.37 & - \\ \bottomrule
\end{tabular}
\end{table*}

\begin{figure*}[t]
    \vspace{5pt}
    \centering
    \includegraphics[width=7in]{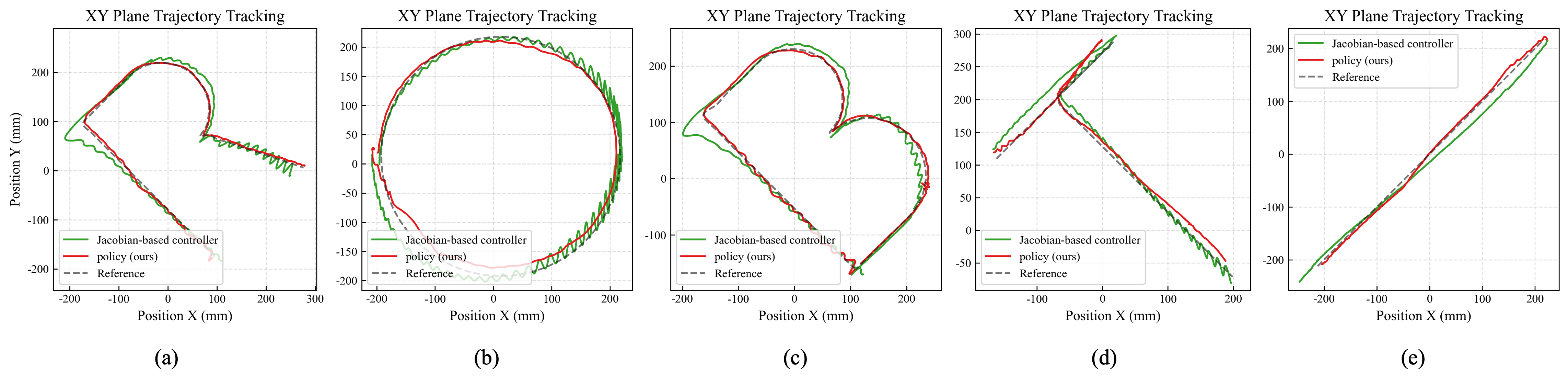}
    \vspace{-20pt}
    \caption{Experimental tip position tracking performance on real-world TDCR at a speed of 23~mm/s. Panels (a) to (e) display the results for the letter-shaped trajectories ``R'', ``O'', ``B'', ``T'' and the straight-line path, respectively. The gray dashed lines represent the reference trajectories. Our proposed policy (red solid line) demonstrates superior tracking precision and smoothness compared to the Jacobian-based controller (green solid line), which exhibits significant deviations and oscillatory behaviors, particularly during sharp turns.}
    \label{fig:xy_traj}
    \vspace{-10pt}
\end{figure*}

\section{Reference Augmentation}
To enhance the robustness of the neural tracking policy and prevent overfitting to specific trajectories in the offline dataset, we implement a multi-scale reference augmentation scheme. This method synthesizes diverse tracking scenarios by injecting structured stochastic perturbations into the ground-truth state sequences during training.

\subsection{Stochastic Noise Components}
For each sampled trajectory starting at time $t$, the augmented reference $\bm{r}_t$ is generated by superimposing three distinct stochastic components. Let $i \in \{1, \dots, N_p\}$ denote the look-ahead horizon. The perturbations are formulated as follows:
\begin{itemize}
    \item {Constant Bias ($\bm{\delta}^{bias}$)}: Captures static coordinate shifts or global goal offsets.
    \begin{equation}
        \bm{\delta}^{bias} \sim \mathcal{U}(-A_b, A_b),
    \end{equation}
    where $\mathcal{U}(\cdot)$ denotes a uniform distribution.

    \item{Harmonic Perturbation ($\bm{\delta}_{sine}$)}: Unlike high-frequency jitter, this component mimics smooth, low-frequency fluctuations over time to simulate dynamic drifts in reference signals. For each relative step $i$:
    \begin{equation}
        \bm{\delta}^{sine}_{t+i} = \bm{A}_s \cdot \sin(2\pi \bm{f} \cdot i + \bm{\phi}),
    \end{equation}
    where the amplitude, frequency, and phase are sampled from: $\bm{A}_s \sim \mathcal{U}(0, A_{s,max})$, $\bm{f} \sim \mathcal{U}(0, f_{max})$, and $\bm{\phi} \sim \mathcal{U}(0, 2\pi)$.

    \item{Random Walk ($\bm{\delta}_{step}$)}: Models cumulative drift and sudden setpoint changes through a masked random walk process:
    \begin{equation}
        \bm{\delta}^{step}_{t+i} = \sum_{k=0}^{i} \bm{m}_k \cdot \bm{w}_k,
    \end{equation}
    where $\bm{m}_k \sim \text{Bernoulli}(p_{step})$ is a binary mask, and $\bm{w}_k \sim \mathcal{U}(-A_w, A_w)$ represents the step magnitude.
\end{itemize}

The total perturbation for a specific trajectory sample starting from time $t$ is formulated as:
\begin{equation}
    \bm{\delta}^{total}_{t+i} = \bm{\delta}^{bias} + \bm{\delta}^{sine}_{t+i} + \bm{\delta}^{step}_{t+i}.
\end{equation}

\subsection{Mixed Reference Strategy}
To decouple the policy's dependency on specific future trajectory shapes and improve its flexibility, we employ a masked mixing strategy. 
For a given ground-truth sequence $\{\bm{T}_{t}, \bm{T}_{t+1}, ..., \bm{T}_{t+N_p}\}$, the base reference $\bm{T}^{base}_{t+i}$ is selected as:
\begin{equation}
    \bm{T}^{base}_{t+i} = M \odot \bm{T}_{t+i} + (\mathbf{1} - M) \odot \bm{T}_t,
\end{equation}
where $M$ is a random binary mask.
This strategy forces the policy to switch between following a time-varying path and maintaining a static setpoint, effectively integrating reaching and tracking tasks into a unified framework.

The final augmented reference is obtained as:
\begin{equation}
    \bm{T}^{*}_{t+i} = \bm{T}^{base}_{t+i} + \bm{\delta}^{total}_{t+i}.
\end{equation}

It is worth noting that we apply these perturbations to the ground-truth future states from the dataset rather than the original reference signals. This ensures that the policy learns to map current observations to perturbed ground-truth states, which forces the policy to learn an implicit recovery mechanism against trajectory deviations, rather than simply memorizing the nominal offline path.

\subsection{Update and Consistency}
During the BPTT rollout, the augmented reference is updated for each sample. To ensure temporal consistency within a single rollout, the noise parameters $(\bm{A}_s, \bm{f}, \bm{\phi})$ and the consistent bias $\bm{\delta}^{bias}$ are initialized at the start of each sequence and kept consistent as the horizon slides. This structured, consistent noise forces the policy to learn compensation for both high-frequency jitter and long-term low-frequency drifts in TDCR operations.

\section{Experiments}
In this section, we evaluate the proposed control policy through three groups of experiments to investigate: (a) the impact of reference augmentation on tracking precision; (b) the performance comparison between our policy and the traditional Jacobian-based controller; and (c) the sensitivity of tracking performance to the look-ahead horizon $N_r$ and the optimization horizon $N_p$.

\subsection{Setup}
The platform consists of a 9-motor tendon-driven continuum robot, a PLC-based actuation system, and a motion capture system as shown in Fig.~\ref{fig:total}(a). The control loop operates at 50~Hz. We utilize a 4-layer RNN of 1024 hidden units, LayerNorm~\cite{ba2016layernormalization}, and 0.3 dropout~\cite{nitish2014dropout} as our policy network. Policy optimization uses an inference horizon $N_i=50$, reference horizon $N_r=50$, and an optimization horizon $N_p=250$ with a discount factor $\gamma=1.0$. All inputs are standardized to unit normal distributions based on the dataset statistics.

Tracking tasks are evaluated at three speeds: 23, 38, and 58~mm/s (corresponding to 1.0$\times$, 1.7$\times$, and 2.5$\times$ scales) using letter-shaped ``R'', ``O'', ``B'', ``T'' and straight-line trajectories.

To ensure diverse and robust training, the reference augmentation parameters are configured based on empirical physical limits. The maximum constant bias $A_b$ is set to 20~mm for position and 0.1~rad for orientation. Harmonic perturbations utilize a maximum frequency $f_{max} = 0.25$~Hz with amplitudes $A_{s,max}$ up to 25~mm and 0.1~rad. The random walk process employs a step rate $p_{step} = 0.001$ with maximum step magnitudes $A_w$ of 20~mm and 0.1~rad for position and orientation, respectively.

\subsection{Comparison Experiments}

The experimental results summarized in Table~\ref{tab:precision_comparison} highlight the significant advantages of the proposed reference augmentation strategy. Compared to the baseline policy trained on raw dataset references without augmentation, our method reduces the average position error by 50\% from 29.00~mm to 14.25~mm and the orientation error by 28.4\% from 8.1\textdegree~to 5.8\textdegree. This substantial performance gain validates that the structured stochastic noise injected during training effectively prevents the model from over-fitting to the limited, nominal trajectories present in the offline dataset. By synthesizing a broader distribution of tracking scenarios, the augmentation scheme forces the policy to learn a more versatile and robust mapping from observations to control actions, enabling it to generalize to unseen motion patterns and system drifts.

Our learning-based policy also consistently outperforms the traditional Jacobian-based controller across all evaluated speeds. As indicated in Table~\ref{tab:precision_comparison}, the Jacobian method struggles with high-dimensional 6-DOF tracking, particularly in orientation, where it exhibits an average error of 22.3\textdegree. In contrast, our policy maintains a stable orientation error of 5.8\textdegree, demonstrating its superior capability in coordinating the complex, coupled degrees of freedom inherent in TDCRs.

Furthermore, we evaluated a \textit{Pos-only} policy to establish the performance upper bound of our framework when focusing solely on Cartesian accuracy. As illustrated in Fig.~\ref{fig:xy_traj}, the Jacobian controller suffers from significant path deviations and self-excited oscillations during sharp maneuvers—behaviors typically triggered by the high nonlinearity and hysteresis of the tendon-driven system. In contrast, our proposed policy achieves high-precision tracking with an average position error of 9.79~mm at 1.0$\times$ speed. The resulting tracking remains remarkably smooth even during complex letter-shaped trajectories. These results underscore that the multi-scale reference augmentation successfully bridges the gap between limited offline data and complex real-world tracking requirements, providing the policy with the necessary exposure to diverse tracking errors to learn effective corrective behaviors.

\begin{figure}[t]
\centering
\includegraphics[width=3.2in]{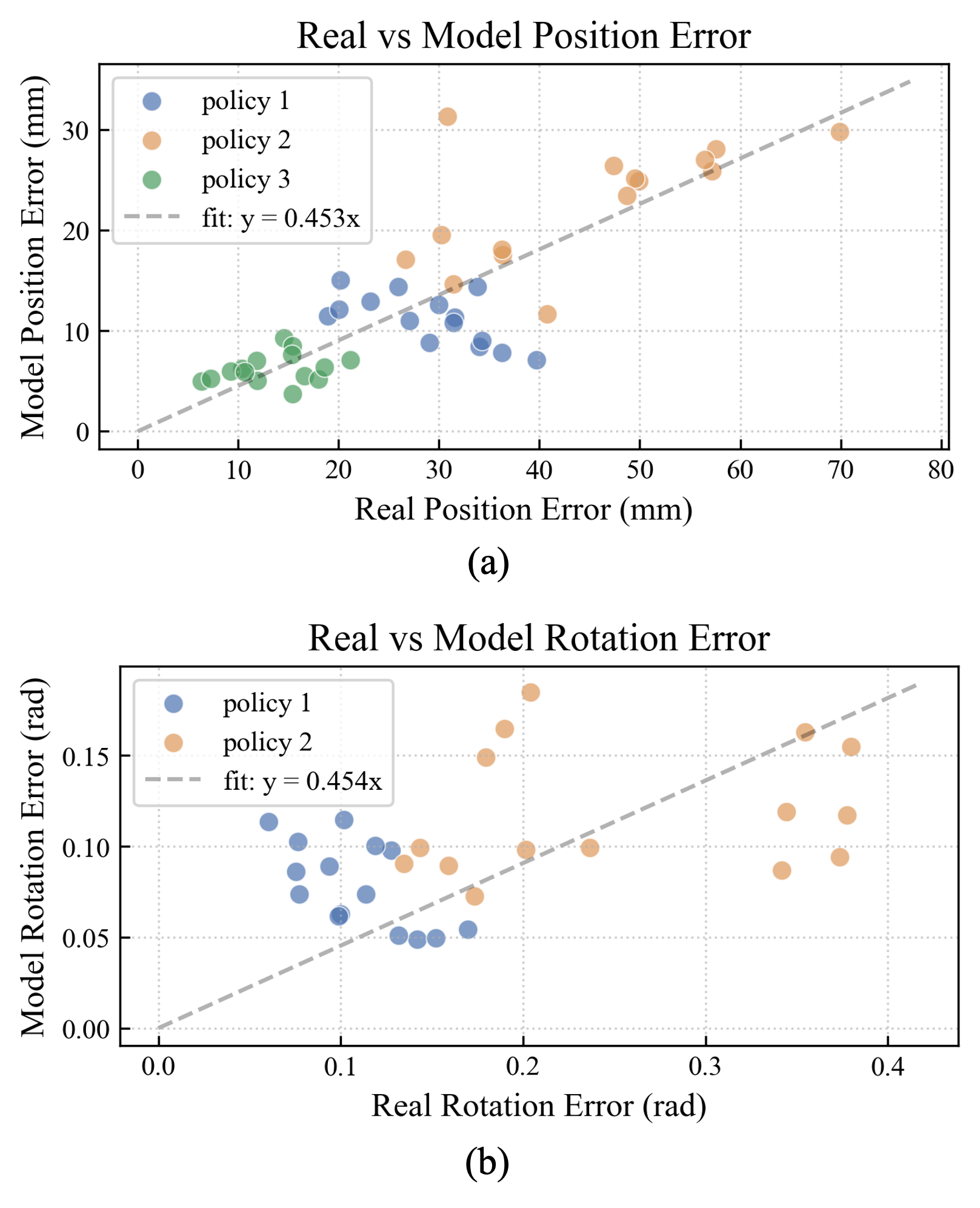}
\vspace{-10pt}
\caption{Correlation analysis of tracking performance between the dynamics surrogate and the real-world TDCR across different policies. (a) Position error and (b) orientation error. The linear fits ($y \approx 0.45x$) indicate that while the surrogate underestimates the absolute error magnitude, it maintains a strong linear correlation with real-world performance.}
\label{fig:model_vs_real}
\vspace{-10pt}
\end{figure}

\subsection{Ablation Study}
While policies with poorly tuned horizons may still achieve mathematical convergence, they often result in aggressive or erratic control actions that could strain the TDCR hardware.
To evaluate the impact of the reference horizon $N_r$ and optimization horizon $N_p$ without risking hardware damage from suboptimal control policies, we utilized the trained dynamics model as a surrogate for full-trajectory rollouts.

\begin{figure}[t]
\centering
\includegraphics[width=3.2in]{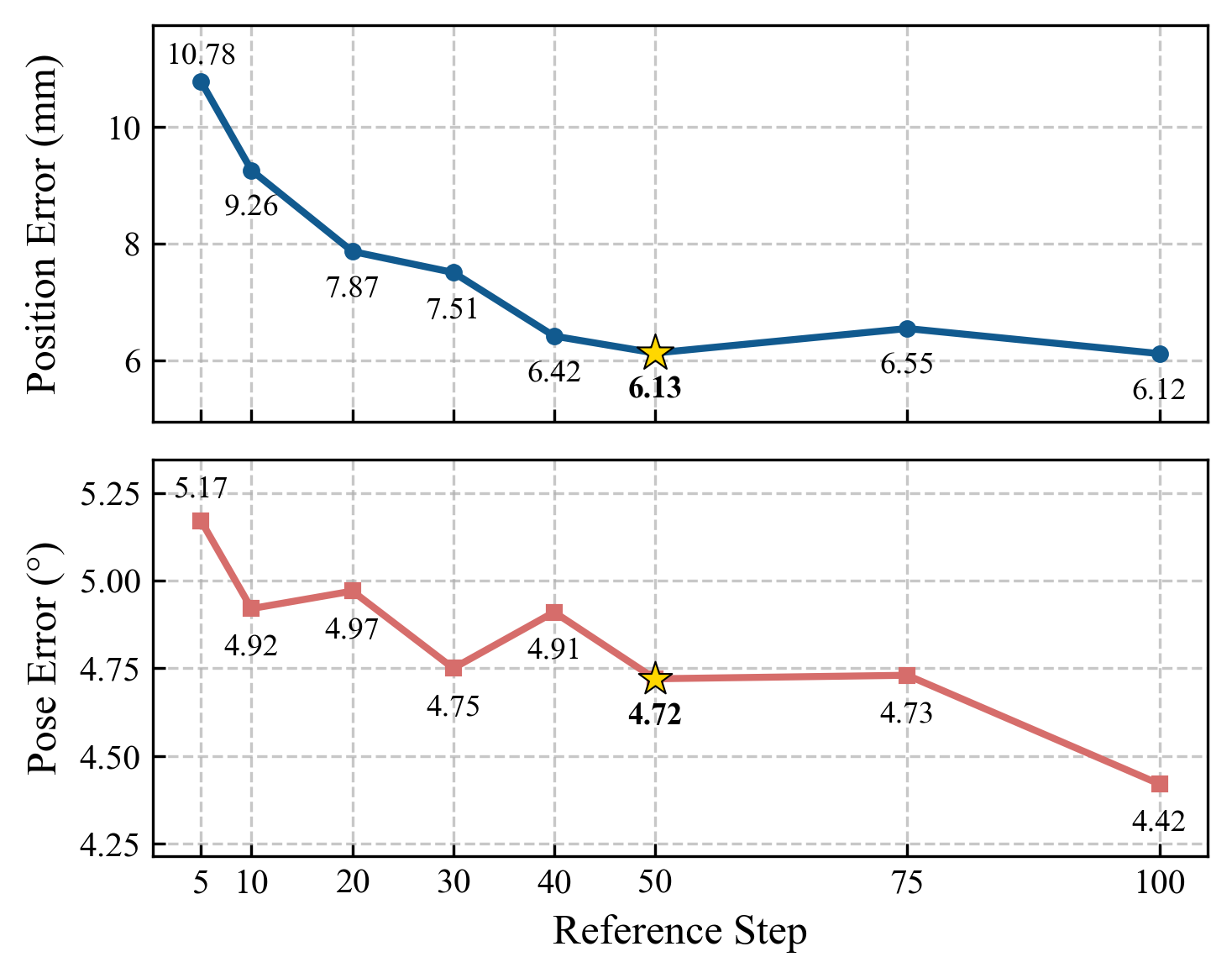}
\vspace{-10pt}
\caption{Sensitivity analysis of the reference horizon $N_r$. Increasing the look-ahead steps provides more future context, leading to a consistent reduction in tracking errors, although marginal returns diminish beyond $N_r=50$.}
\label{fig:ref_step}
\end{figure}

\begin{figure}[t]
\centering
\includegraphics[width=3.2in]{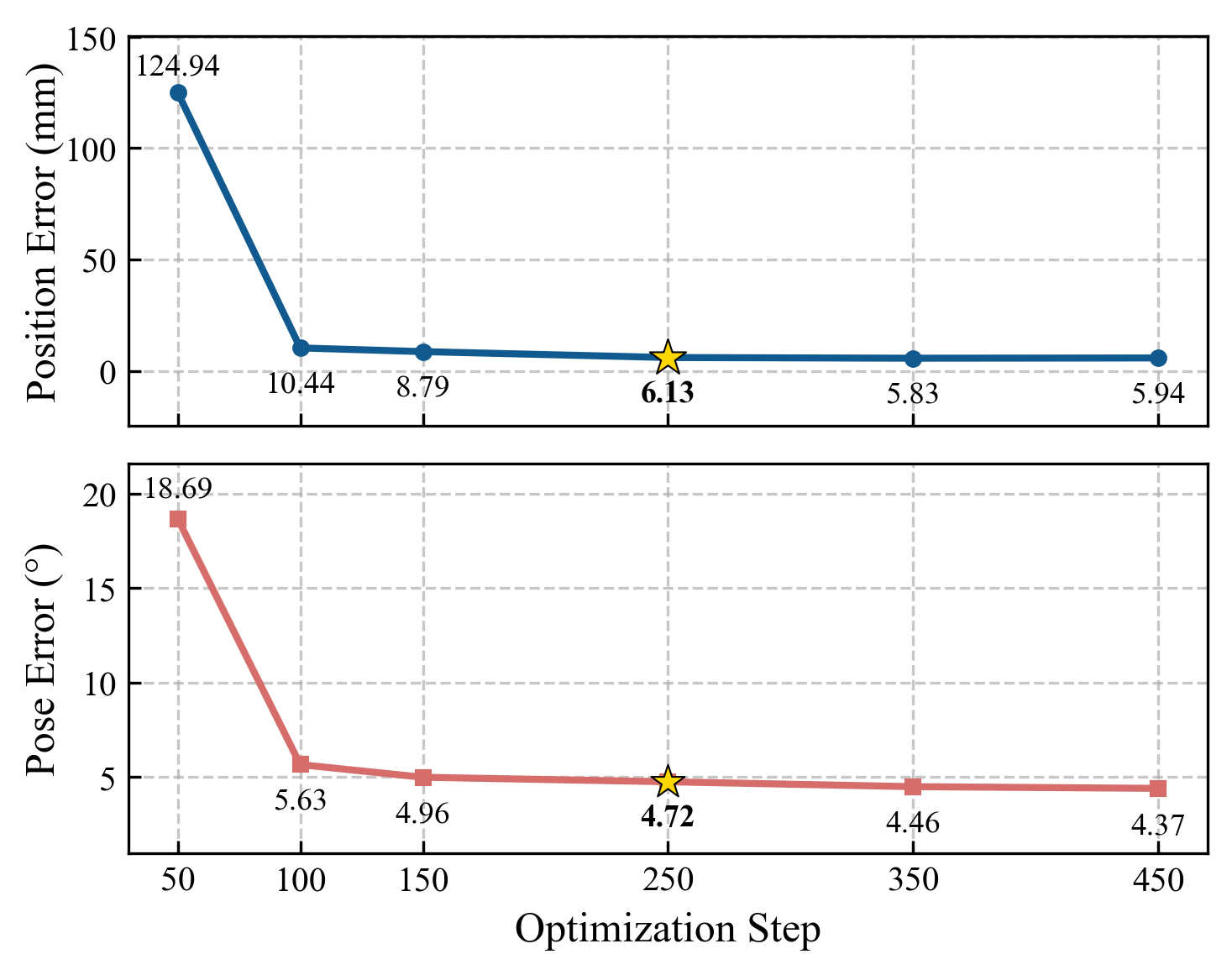}
\vspace{-10pt}
\caption{Sensitivity analysis of the optimization horizon $N_p$. Tracking performance suffers significantly when $N_p$ is too short, with convergence typically requiring a horizon of at least 100 steps.}
\label{fig:opt_step}
\end{figure}

\addtolength{\textheight}{-2cm}   

As illustrated in Fig.~\ref{fig:model_vs_real}, we conducted a correlation analysis by evaluating various policies on both the physical robot and the dynamics surrogate. The results show a strong positive correlation between the tracking errors observed in the surrogate environment and those on the real-world platform for both position and orientation. This consistent trend validates the feasibility of using the dynamics surrogate as a reliable proxy for safety-aware policy ablation.

To explore the influence of key temporal parameters on tracking performance, we conducted a sensitivity analysis on the reference horizon $N_r$ and the optimization horizon $N_p$.

As illustrated in Fig.~\ref{fig:ref_step}, increasing the reference horizon $N_r$ generally enhances precision by providing the policy with broader future context. However, a clear law of diminishing marginal utility is observed. The error reduction becomes significantly less pronounced once $N_r$ exceeds 50 steps, suggesting that a 1-second look-ahead is sufficient for most TDCR tracking maneuvers.

More importantly, Fig.~\ref{fig:opt_step} demonstrates that tracking stability is highly sensitive to the optimization horizon $N_p$. Specifically, at $N_p=50$, corresponding to a 1~s window, the tracking fails with a drastic position error of 124.94~mm. This failure is likely due to the significant hysteresis and structural compliance inherent in TDCRs, which require a sufficiently long control horizon to allow the system dynamics to reach a steady state. Our results indicate that while further increasing $N_p$ beyond 150 steps yields slight improvements. To strike a balance between tracking precision and computational overhead, we select $N_p=250$ as the standard configuration for our control policy.

\section{Conclusion}
In this paper, we presented a reference-augmented learning framework to address the precise trajectory tracking problem for TDCRs. By integrating a differentiable dynamics surrogate with a structured reference randomization strategy, our approach effectively bridges the gap between limited offline datasets and complex real-world operational requirements. Experimental validations on a physical platform confirm that the proposed policy consistently achieves high-precision 6-DOF tracking, maintaining an average position error of 14.25 mm and orientation error of 5.8\textdegree~even under high-speed maneuvers where traditional Jacobian-based controllers exhibit significant oscillations. Sensitivity analyses further underscore the critical role of long optimization horizons in compensating for the inherent hysteresis and compliance of TDCRs. The results demonstrate that our framework provides a robust and sample-efficient solution for the control of soft robotic systems, offering a scalable path toward autonomous dexterous manipulation in constrained environments.


\bibliographystyle{IEEEtran}
\bibliography{ref}

\end{document}